\documentclass[conference]{IEEEtran}
\IEEEoverridecommandlockouts

\usepackage{graphicx}
\usepackage{amsmath,amssymb}
\usepackage{cite}
\usepackage{subcaption}
\usepackage{url}
\usepackage{siunitx}   
\usepackage{booktabs}  
\usepackage{hyperref} 
\usepackage{algorithm}
\usepackage[noend]{algpseudocode}
\begin{document}

\title{Hybrid Terrain-Aware Path Planning: Integrating VD--RRT\(^{*}\) Exploration and VD--D\(^{*}\) Lite Repair}

\author{Akshay Naik$^{1}$,
        William~R.~Norris$^{2\dag}$,
        Dustin~Nottage$^{3}$,
        Ahmet~Soylemezoglu$^{3}$

\thanks{Akshay Naik$^{1}$ is with The Grainger College of Engineering, Electrical and Computer Engineering Department, University of Illinois Urbana-Champaign, Urbana, IL 61801-3080 USA (email: akshayn3@illinois.edu).}%

\thanks{William R.~Norris$^{2\dag}$ was with The Grainger College of Engineering, Industrial and Enterprise Systems Engineering Department, University of Illinois Urbana-Champaign, Urbana, IL 61801-3080 USA. \dag\emph{Deceased.}}%

\thanks{Construction Engineering Research Laboratory$^{3}$, U.S. Army Corps of Engineers Engineering Research and Development Center, IL, 61822, USA.}%

\thanks{This research was supported by the U.S. Army Corps of Engineers Engineering Research and Development Center, Construction Engineering Research Laboratory.}%
}

\maketitle
\begin{abstract}
Autonomous ground vehicles operating off-road must plan curvature-feasible paths while accounting for spatially varying soil strength and slope hazards in real time. We present a continuous state--cost metric that combines a Bekker pressure--sinkage model with elevation-derived slope and attitude penalties. The resulting terrain cost field is analytic, bounded, and monotonic in soil modulus and slope, ensuring well-posed discretization and stable updates under sensor noise. This metric is evaluated on a lattice with exact steering primitives: Dubins and Reeds--Shepp motions for differential drive and time-parameterized bicycle arcs for Ackermann steering. Global exploration is performed using Vehicle-Dynamics RRT\(^{*}\), while local repair is managed by Vehicle-Dynamics D\(^{*}\) Lite, enabling millisecond-scale replanning without heuristic smoothing. By separating the terrain--vehicle model from the planner, the framework provides a reusable basis for deterministic, sampling-based, or learning-driven planning in deformable terrain. Hardware trials on an off-road platform demonstrate real-time navigation across soft soil and slope transitions, supporting reliable autonomy in unstructured environments.
\end{abstract}


\section{Introduction}
\label{sec:intro}
Autonomous ground vehicles operating beyond paved roads must navigate soils that deform, slopes that destabilize, and vegetation that obscures traversable space. In such environments, terrain can shift from passable to impassable within a few meters, forcing planners to react in real time. The traversability of a cell is therefore not binary but a continuous function of sinkage, slip, and rollover risk. Vehicles may also face different kinematic limits—such as skid-steer rovers capable of turning in place or Ackermann tractors constrained by minimum curvature—so any path that ignores vehicle dynamics often fails at execution. Bridging terrain physics with non-holonomic motion while still enabling real-time replanning remains a central challenge for off-road autonomy.

This work proposes a terrain-aware planning framework that unifies soil mechanics, slope effects, and vehicle kinematics under a single analytic cost field and curvature-constrained lattice. Specifically, we: (i) derive a bounded, monotone cost metric from Bekker pressure–sinkage theory augmented with slope and attitude penalties; (ii) embed exact Dubins, Reeds--Shepp, and bicycle primitives into a lattice that guarantees curvature feasibility at graph construction; and (iii) combine Vehicle-Dynamics RRT\(^{*}\) for global refinement with Vehicle-Dynamics D\(^{*}\) Lite for millisecond-scale local repair. Together, these components form a practical, reproducible pipeline that enables consistent planning across kilometer-scale terrain while adapting automatically to soil and elevation updates.

Research to date has advanced along two largely separate tracks. Grid-based algorithms such as A* and D\(^{*}\) Lite deliver deterministic shortest paths and react quickly to local cost changes \cite{lavalle_planning_2006,koenig_dlite_2002}, but typically encode motion using four- or eight-connected neighbors and model cost as Euclidean distance or inflated occupancy, leaving soil strength and curvature unrepresented. Sampling-based planners like RRT\(^{*}\) explore the continuous configuration space and incorporate steering primitives, yet assume static cost maps and incur significant overhead when obstacles appear or disappear \cite{karaman_sampling-based_2011,lavalle_planning_2006}. Dynamic and informed extensions (RRTX, BIT*) reduce this overhead but still neglect rolling resistance and terrain-dependent cost \cite{otte_rrtx_2016,gammell_batch_2015}; bidirectional informed variants further improve convergence in clutter \cite{fan_bi-rrt_2024}. Hybrid schemes attach curvature-constrained primitives to grids or use lazy repair on RRT trees \cite{huang_asymptotically_2024}, but terrain variability is often compressed into a single heuristic weight that overlooks soil mechanics \cite{huang_asymptotically_2024,zhang_motion_2025}.  

By contrast, our framework provides a generic, physics-grounded abstraction of terrain that existing planners can query directly. Soil sinkage, slip, and slope are all incorporated into the cost field, and curvature feasibility is enforced natively through the lattice. This approach allows unforeseen terrain variations to be handled in real time without ad hoc rules, making the system suitable for real-world deployment. Although evaluation in this paper focuses on synthetic maps and preliminary field tests, the framework is currently implemented on a research platform and being extended for large-scale experiments.

\begin{figure}
    \centering
    \includegraphics[width=\linewidth]{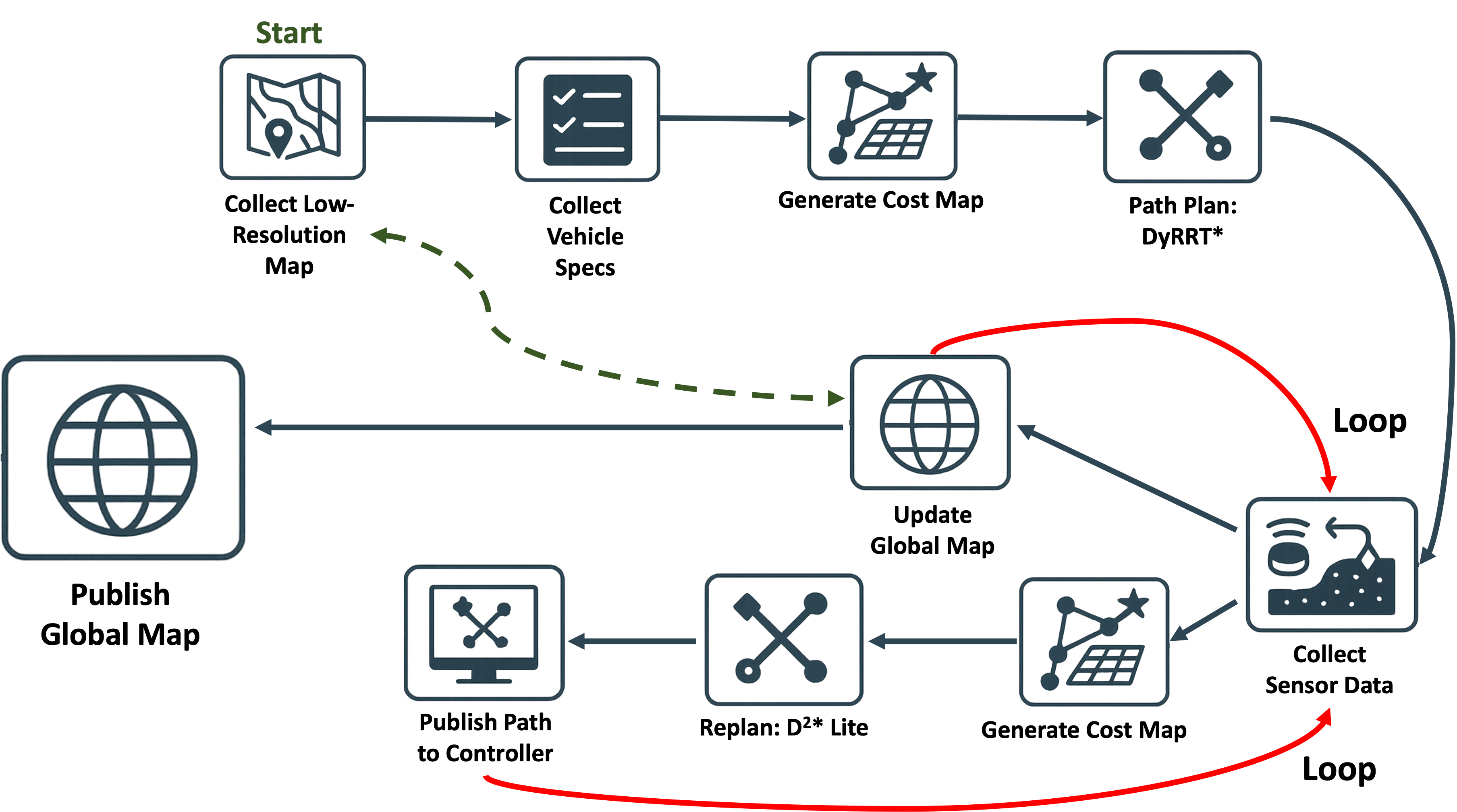}
    \caption{End-to-end planning pipeline coupling a static VD--RRT$^{*}$ seed with real-time VD--D\(^{*}\) Lite repair.}
    \label{fig:pipeline_overview}
\end{figure}

To exploit the lattice, we combine two complementary search routines. Vehicle-Dynamics RRT\(^{*}\) grows a low-dispersion tree that converges toward the global optimum, while Vehicle-Dynamics D\(^{*}\) Lite repairs affected vertices when LiDAR scans or soil updates alter only a few cells. Edge validity depends solely on the updated cost raster, so new terrain information is automatically incorporated. The result is a planner that fuses steady global refinement with real-time local correction, producing safe, consistent paths in variable terrain, as summarized in the end-to-end pipeline of Figure~\ref{fig:pipeline_overview}.

The remainder of the paper is organized as follows. Section II formalizes the Bekker--slope cost and its analytic properties. Section III presents the full methodology, including the curvature-constrained lattice, hybrid VD--RRT\(^{*}\) / VD--D$^{*}$ Lite planner, and hardware validation pipeline. Section IV reports results and analysis on both synthetic benchmarks and off-road trials. Section V provides discussion of limitations and broader implications. Section VI concludes with directions for future work. By unifying terramechanics with non-holonomic motion in a planner-agnostic abstraction, we aim to advance physics-aware autonomy for off-road environments.

\section{Related Work}
\label{sec:related_work}
Classical planners such as A* and Dijkstra remain foundational for robot navigation~\cite{lavalle_planning_2006}, but their binary free--occupied abstraction collapses traversability into geometry and ignores vehicle curvature limits. Incremental search methods such as D$^{*}$ Lite~\cite{koenig_dlite_2002} reduce replanning cost by reusing prior search effort, yet their costs remain largely Euclidean and lack mechanisms to penalize soft soil or steep slopes.  

Sampling-based planners address kinematic feasibility by exploring continuous state spaces. RRT and PRM families~\cite{lavalle_planning_2006} yield feasible trajectories in high dimensions, with asymptotic optimality introduced by RRT$^{*}$~\cite{karaman_sampling-based_2011}. Dynamic extensions such as RRTX~\cite{otte_rrtx_2016} and BIT*~\cite{gammell_batch_2015} improve solution quality while reacting to changes, but their edge costs are still geometric. Curvature-feasible primitives have been studied for car-like and heavy-duty vehicles~\cite{reeds_optimal_1990,oliveira_trajectory_2018,hoffmann_optimal_2023}, yet most implementations assume rigid terrain and neglect soil mechanics.  

Recent work has explored terrain-aware navigation. Classical terramechanics~\cite{bekker_introduction_1969,wong_theory_2008,wong_prediction_1967} relates wheel load and sinkage, but robotic planners often substitute heuristics such as slope or roughness indices~\cite{gu_rapid_2008}. More recent traversability mapping fuses vision and LiDAR to classify terrain~\cite{zhou_terrain_2022,guan_tnes_2023}, or applies deep learning to predict pixel-level costs from images and point clouds~\cite{lin_accurate_2024,dong_superfusion_2024}. While these systems provide semantic cues, they require large annotated datasets and ultimately rely on hand-tuned mappings from labels to planner costs.  

Our approach differs by embedding an analytic Bekker-derived soil term and slope/attitude penalties directly into a curvature-feasible lattice. Unlike fusion pipelines that output categorical traversability~\cite{zhou_terrain_2022,liu_real_2023}, our cost is continuous, monotone, and physics-grounded. We further couple a global VD--RRT\(^{*}\) module with a fast VD--D$^{*}$ Lite repairer, reusing edge evaluations for millisecond-scale replanning. This hybridization closes the gap between geometry-only incremental planners and computationally expensive terramechanics simulators, enabling reproducible real-time navigation in deformable outdoor settings.

\section{Methodology}
\label{sec:methodology}
Planning on deformable ground requires a wheel–soil interaction model that can be evaluated thousands of times per second without resorting to heavy elastoplastic continuum mechanics. Among the available formulations, we adopt Bekker’s semi-empirical pressure–sinkage law as it offers an attractive compromise between fidelity and efficiency. Originally derived by Bekker~\cite{bekker_introduction_1969} and validated across lunar simulants, sand, loam, and gravel~\cite{wong_prediction_1967}, the model ties vertical pressure under a wheel directly to sinkage via a power law with three parameters: \(k_c\) (cohesive modulus), \(k_{\phi}\) (frictional modulus), and \(n\) (sinkage exponent). These parameters can be estimated rapidly from plate-sinkage or cone-index tests, making them practical for deployment.  

We embed this soil model, together with slope and attitude penalties, into a curvature-feasible lattice whose edges are exact steering primitives. Global exploration is performed once at initialization with Vehicle-Dynamics RRT\(^{*}\), using DEM priors or coarse maps to supply long-horizon structure. During execution, Vehicle-Dynamics D$^{*}$ Lite incrementally repairs the same lattice as live sensor costs arrive, reusing edges whenever possible. This hybrid ensures that the global seed and local repairs share a single physics-grounded substrate, yielding paths that are both globally consistent and locally reactive while remaining lightweight enough for CPU-only embedded hardware.

\begin{figure}[t!]
  \centering
  \includegraphics[width=0.75\linewidth]{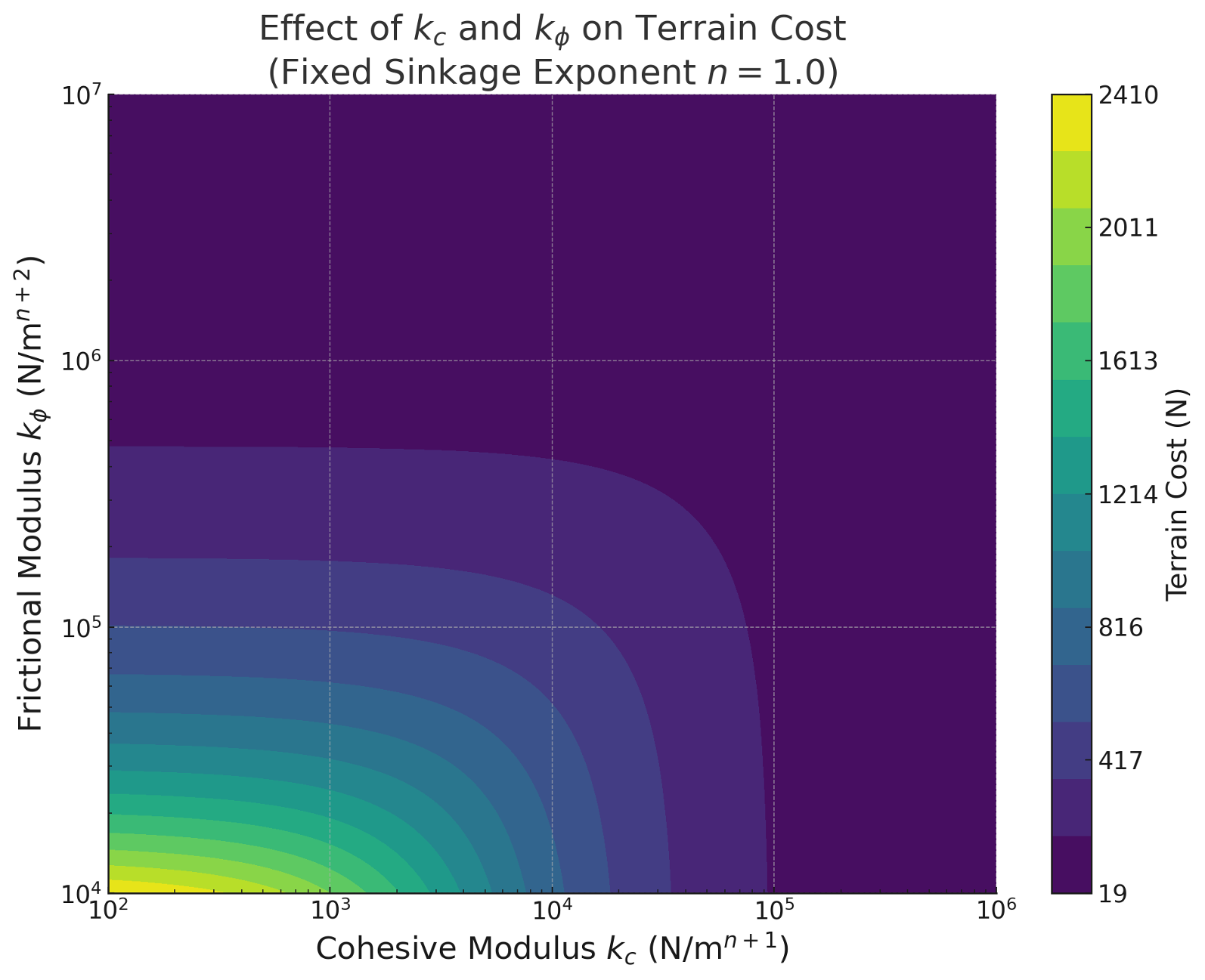}
  \caption{Terrain cost as a function of cohesive modulus ($k_c$) and frictional modulus ($k_\phi$). Softer soils (low $k_c$, $k_\phi$) correspond to higher traversal cost.}
  \label{fig:bekker_surface}
\end{figure}

Figure~\ref{fig:bekker_surface} illustrates how the terrain cost varies with the two Bekker moduli. The pressure–sinkage law is given by
\[
  p(z) = \Bigl(\tfrac{k_c}{b} + k_{\phi}\Bigr)\,z^{\,n},
\]
where \(z\) is sinkage depth and \(b\) is the smaller contact-patch dimension. Here, \(k_c\) and \(k_{\phi}\) are analogous to the intercept and slope of a Mohr–Coulomb envelope, while \(n\) captures soil compressibility (\(n\!\approx\!1.2\) for sand, \(n\!\approx\!0.6\) for clay). All quantities are SI; $b$ is in meters, and $(k_c,k_\phi,n)$ carry units that ensure $p(z)$ has units of pressure.

For a wheel of radius \(R\) carrying per-wheel load \(W_{\text{contact}}\), we approximate the contact area as \(A \simeq b\sqrt{2Rz}\). By applying static equilibrium \(W_{\text{contact}} = p(z)A\), we obtain
\[
  z = \Biggl[
        \frac{W_{\text{contact}}}
             {(\tfrac{k_c}{b}+k_{\phi})\,b\,\sqrt{2R}}
      \Biggr]^{\frac{1}{n+\tfrac12}}.
\]
This expression shows that larger wheels or stiffer soils reduce sinkage, while heavier loads increase it.

\medskip
\noindent\textbf{Soil traversal cost.}
To make terrain conditions directly queryable by the planner, we adopt Bekker’s classical pressure--sinkage law~\cite{bekker_introduction_1969} and normalize sinkage depth by wheel radius to obtain a dimensionless soil cost:
\[
C_{\text{soil}}
= \frac{z}{R}
= \frac{1}{R}
  \left[
      \frac{W_{\text{contact}}}
           {(\tfrac{k_c}{b}+k_{\phi})\,b\,\sqrt{2R}}
  \right]^{\tfrac{1}{n+\tfrac12}},
\]
where $z$ is the predicted vertical sinkage, $W_{\text{contact}} = W_{\text{total}}/N$ is the per-wheel load, $R$ is wheel radius, and $b$ is tire width. This form is unitless, grows monotonically with sinkage risk, and is naturally bounded by clipping $C_{\text{soil}}\leq1$, corresponding to full wheel burial. Candidate edges are evaluated by integrating $C_{\text{soil}}$ along their footprint, so paths naturally steer clear of soft-soil corridors.

\medskip
\noindent Table~\ref{tab:bekker_coeffs_verbose} lists nominal Bekker parameters $(k_c, k_\phi, n)$ for several common outdoor substrates. These values are representative engineering estimates consistent with ranges reported in terramechanics literature~\cite{bekker_introduction_1969,wong_prediction_1967,wong_theory_2008}. They are not site-specific measurements but serve as practical lookup entries in our pipeline. In simulation, synthetic maps embed terrain labels that index directly into the table; in hardware, the vision-based soil classifier outputs the same labels, which are mapped to the same coefficients. This ensures that both environments rely on a consistent, physics-informed soil cost model without requiring ad hoc tuning.

\begin{table}[ht]
  \centering
  \caption{Representative Bekker parameters used as lookup values in both simulation and hardware pipelines. 
  Values are engineering estimates consistent with ranges in terramechanics literature~\cite{bekker_introduction_1969,wong_prediction_1967,wong_theory_2008}. 
  A uniform $\pm20\%$ margin is applied during planning to reflect natural variability.}
  \label{tab:bekker_coeffs_verbose}
  \sisetup{scientific-notation=true, round-mode=figures, round-precision=2}
  \begin{tabular}{@{}l S[table-format=1.1e1] S[table-format=1.1e1] S[table-format=1.1]@{}}
    \toprule
    Soil type & {$k_c$ [N\,m$^{-(n+1)}$]} & {$k_{\phi}$ [N\,m$^{-(n+2)}$]} & {$n$} \\
    \midrule
    Pavement (compacted) & 1.0e6 & 1.0e7 & 1.0 \\
    Gravel (3--6\,mm)    & 0     & 5.0e5 & 1.0 \\
    Wood chips (dry)     & 7.0e3 & 1.5e6 & 0.8 \\
    Loam / field dirt    & 1.0e3 & 1.8e6 & 1.0 \\
    Grass (moist)        & 1.0e3 & 1.2e6 & 0.9 \\
    Loose dune sand      & 2.0e3 & 5.0e5 & 1.2 \\
    \bottomrule
  \end{tabular}
\end{table}

\medskip
\noindent To capture variability from moisture, compaction, and organic content, we propagate a $\pm20\%$ margin through $C_{\text{soil}}$. Both mean and upper-confidence-bound maps are retained, with the latter used for fail-safe planning.

\medskip
\noindent\textbf{Slope and attitude costs.}
In addition to soil mechanics, terrain slope and vehicle attitude strongly affect traversability. Let \(h:\mathbb{R}^2\!\to\!\mathbb{R}\) denote the DEM height (m); its gradient
\(\nabla h = [\partial h/\partial x,\ \partial h/\partial y]^\top\) is evaluated with centered finite differences on a grid of cell size $\Delta$. We use the slope magnitude and saturate it at a threshold \(\tau_{\text{slope}}\):
\[
  C_{\text{slope}}(x,y) = \min\!\big(\,\|\nabla h(x,y)\|,\ \tau_{\text{slope}}\,\big).
\]
To account for rollover, we settle all four tire contact points onto the DEM, compute pitch \(\alpha\) and roll \(\beta\), and penalize exceedance using
\[
  C_{\text{att}}(\alpha,\beta)=
  \max\!\Bigl(0,\tfrac{|\alpha|-\alpha_{\max}}{\alpha_{\max}}\Bigr)
  + \max\!\Bigl(0,\tfrac{|\beta|-\beta_{\max}}{\beta_{\max}}\Bigr).
\]

\medskip
\noindent\textbf{Total edge cost.}
We combine these three components into an integrated cost functional:
\[
  J(e) = \int_{e}\!\Bigl[\,1 
  + \lambda_{\text{slope}}\,C_{\text{slope}}(s)
  + \lambda_{\text{soil}}\,C_{\text{soil}}(s)
  + \lambda_{\text{att}}\,C_{\text{att}}(s)\Bigr]\,\mathrm{d}s.
\]
Here the leading 1 recovers Euclidean length. For sampling and visualization, we also form a fused raster $C_{\text{total}} = \alpha\,C_{\text{slope}} + C_{\text{soil}}$ with $\alpha=1.5$ fixed in all experiments. This formulation is \emph{plug-and-play}: any planner can query the same $J(e)$ without modification.
\begin{figure}[t]
  \centering
  \includegraphics[width=\linewidth]{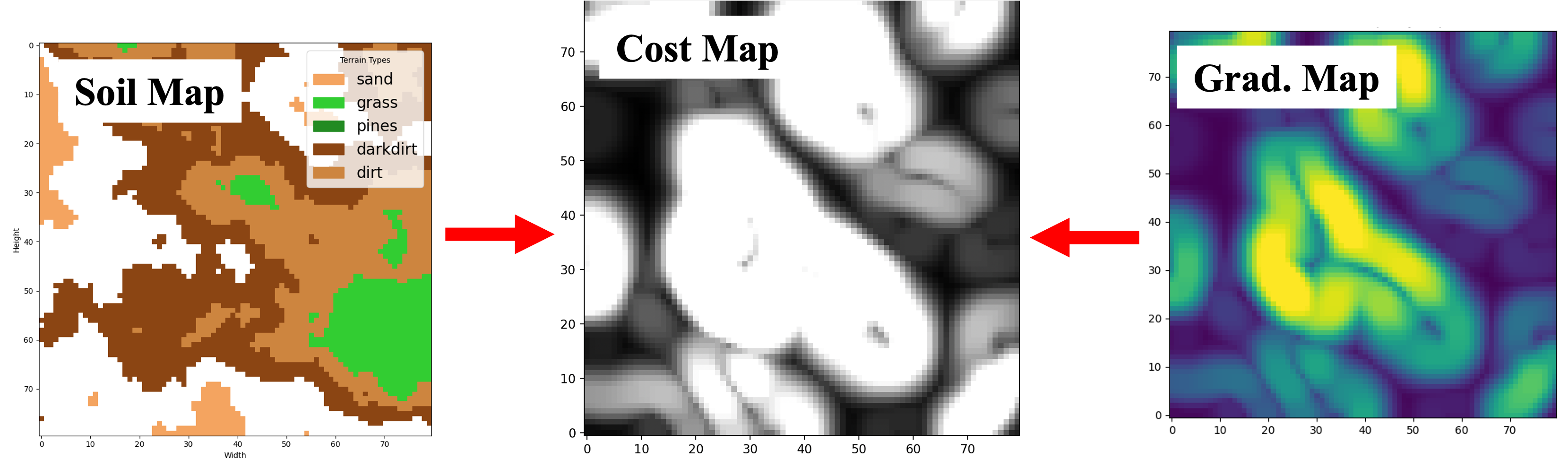}
  \caption{Cost map construction: soil (top-left) and slope (top-right) fuse into the combined raster $C_{\text{total}}$ (bottom).}
  \label{fig:costmap_fusion}
\end{figure}

\begin{figure}[t]
  \centering
  \begin{subfigure}[b]{0.48\linewidth}
    \includegraphics[width=\linewidth]{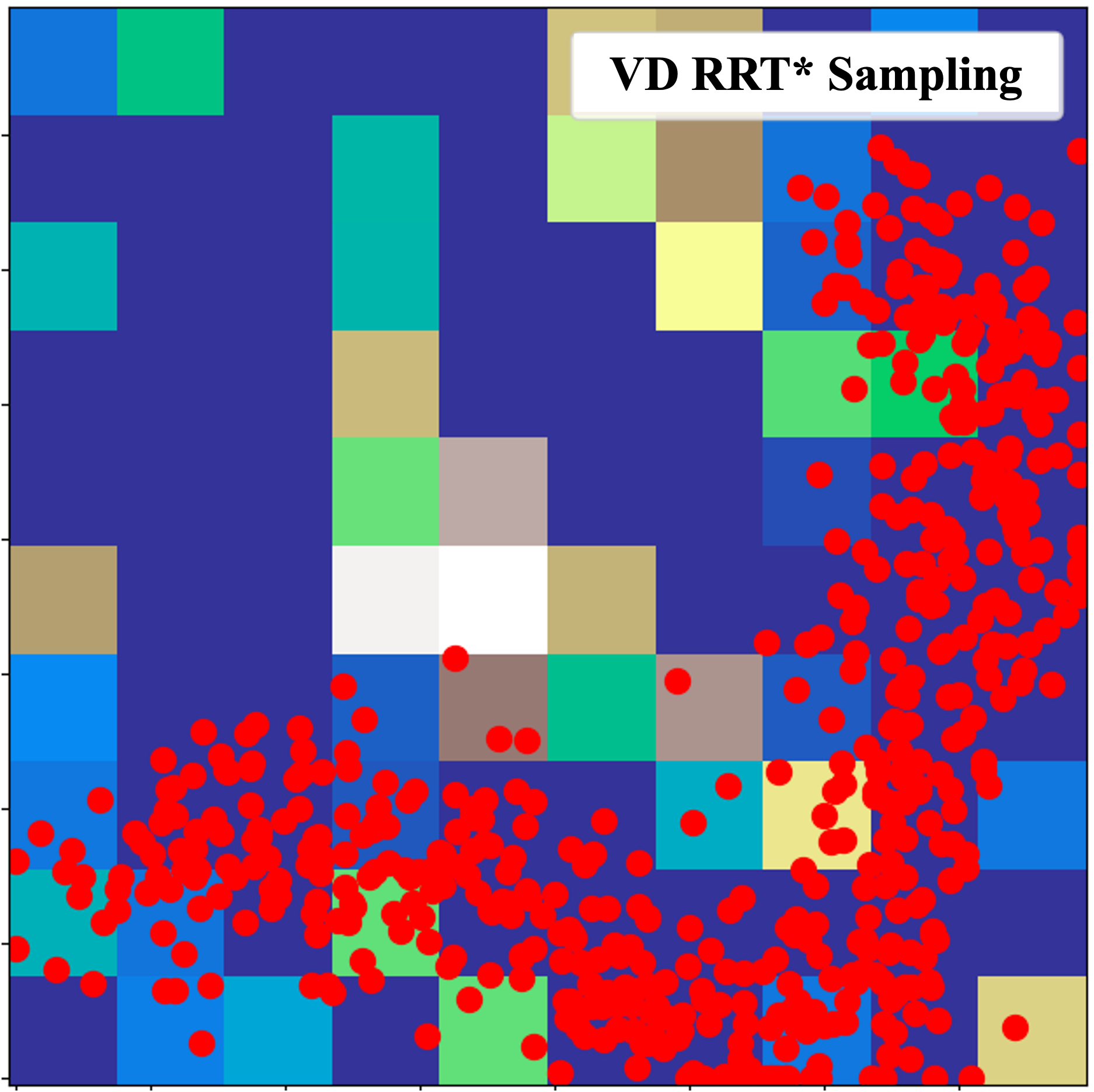}
    \caption{Biased sampling}
  \end{subfigure}
  \hfill
  \begin{subfigure}[b]{0.48\linewidth}
    \includegraphics[width=\linewidth]{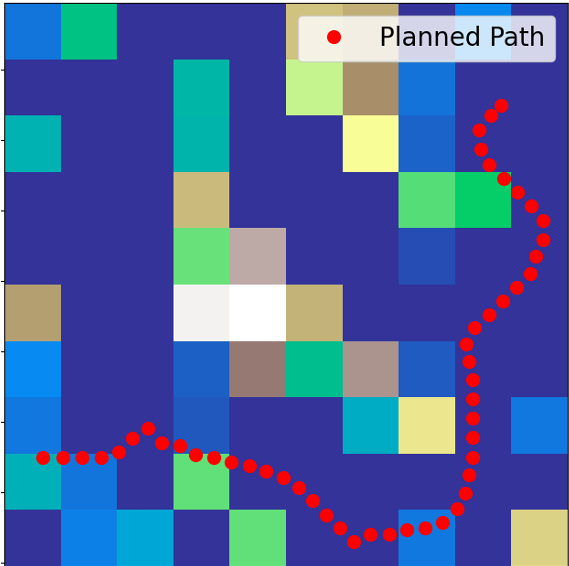}
    \caption{Extracted path}
  \end{subfigure}
  \caption{VD--RRT$^{*}$: samples concentrate in low-cost regions; the final path is curvature-feasible and avoids soft soil.}
  \label{fig:rrt_vis}
\end{figure}

\medskip
\noindent\textbf{Curvature-feasible lattice and VD--RRT\(^{*}\).}
We embed these costs into a curvature-feasible lattice (Figure~\ref{fig:costmap_fusion}). Each state is \((x,y,\theta,v,\sigma)\), where \(\sigma \in \{\mathrm{diff},\mathrm{ack}\}\) distinguishes drive type. Orientation is quantized (16–32 bins) to bound branching. For differential drive, we use Dubins/Reeds--Shepp primitives \cite{reeds_optimal_1990}; for Ackermann steering, we integrate the bicycle model with bounded steering angle. Edges are admissible if all rasterized footprint cells remain below $\mathcal{C}_{\text{high}}$ on $C_{\text{total}}$, and their cost is $J(e)$ via Gauss–Legendre quadrature. Sampling is biased toward low-cost cells (roulette on $1/(1+C_{\text{total}})$) with 5\% goal bias. Nearest-neighbor queries use a k-d tree, and rewiring follows $r(n)\propto\sqrt{\log n/n}$ (Figure~\ref{fig:rrt_vis}).
Algorithm~\ref{alg:veh_dyn_rrtstar} summarizes VD--RRT\(^{*}\).
\begin{algorithm}[t]
  \caption{\textsc{VD--RRT\(^{*}\)} with curvature-feasible primitives and terrain-aware cost}
  \label{alg:veh_dyn_rrtstar}
  \begin{algorithmic}[1]
    \Require start $\mathbf{x}_0$, goal region $\mathcal{X}_{\text{goal}}$, drive mode $\sigma^*$,
             vehicle params $(w,L,R_{\min},\phi_{\max})$, cost rasters $(C_{\text{soil}},C_{\text{slope}},C_{\text{att}})$
    \State $T\gets\{\mathbf{x}_0\}$; initialize k-d tree
    \For{$k=1$ \textbf{to} $K_{\max}$}
      \State $\mathbf{x}_{\mathrm{rand}}\gets\textsc{SampleRoulette}(1/(1+C_{\text{total}}),0.05)$
      \State $\mathbf{x}_{\mathrm{near}}\gets\textsc{Nearest}(T,\mathbf{x}_{\mathrm{rand}})$
      \State $\mathbf{x}_{\mathrm{new}}, e \gets \textsc{SteerPrimitive}(\mathbf{x}_{\mathrm{near}},\mathbf{x}_{\mathrm{rand}},\sigma^*)$
      \If{\textsc{Admissible}$(e,\mathcal{C}_{\text{high}})$}
        \State $c\gets \textsc{IntegrateCost}(J,e)$
        \State $T.\textsc{InsertNode}(\mathbf{x}_{\mathrm{new}},\mathbf{x}_{\mathrm{near}},c)$
        \ForAll{$\mathbf{x}\in \textsc{Near}(T,\mathbf{x}_{\mathrm{new}}, r(n))$}
          \If{\textsc{Admissible}$(\mathbf{x}\to\mathbf{x}_{\mathrm{new}})$}
            \State rewire if new cost improves
          \EndIf
        \EndFor
      \EndIf
      \If{$\mathbf{x}_{\mathrm{new}}\in\mathcal{X}_{\text{goal}}$}
        \State \Return $\textsc{ExtractPath}(T,\mathbf{x}_{\mathrm{new}})$
      \EndIf
    \EndFor
    \State \Return \textsc{Failure}
  \end{algorithmic}
\end{algorithm}

\medskip
\noindent\textbf{Incremental repair with Vehicle-Dynamics D\(^{*}\) Lite.}
For local replanning, we adapt D\(^{*}\) Lite to the same lattice. Successors are generated from the identical motion primitives, and edge weights are computed with $J(e)$. Repairs are triggered when a committed path cell exceeds $\mathcal{C}_{\text{high}}$ or when total path cost increases beyond $\epsilon$. Complexity remains $\mathcal{O}(|E|\log|V|)$, up to a constant for the two steering modes. We use a consistent heuristic $h(s)$ equal to Euclidean distance to the goal multiplied by the unit length cost. Algorithm~\ref{alg:veh_dyn_d2lite} summarizes the procedure.
\begin{algorithm}[t]
  \caption{\textsc{VD--D\(^{*}\) Lite}: incremental repair with terrain-aware edge costs}
  \label{alg:veh_dyn_d2lite}
  \begin{algorithmic}[1]
    \Require start $s_{\mathrm{start}}$, goal $s_{\mathrm{goal}}$, drive mode $\sigma^*$
    \State initialize $g(s)\gets\infty$, $rhs(s)\gets\infty$; $rhs(s_{\mathrm{goal}})\gets0$
    \State insert $s_{\mathrm{goal}}$ in OPEN with key $k(s_{\mathrm{goal}})$
    \While{OPEN.topKey $<$ Key($s_{\mathrm{start}}$) \textbf{ or } $g(s_{\mathrm{start}})\neq rhs(s_{\mathrm{start}})$}
      \State $u \gets$ \textsc{Pop}(OPEN)
      \If{$g(u) > rhs(u)$} \Comment{over-consistent}
        \State $g(u) \gets rhs(u)$
        \ForAll{$s \in \mathrm{Pred}(u)$}
          \State \Call{UpdateVertex}{$s$}
        \EndFor
      \Else \Comment{under-consistent}
        \State $g(u) \gets \infty$
        \ForAll{$s \in \mathrm{Pred}(u)\cup\{u\}$}
          \State \Call{UpdateVertex}{$s$}
        \EndFor
      \EndIf
    \EndWhile

    \Function{UpdateVertex}{$s$}
      \If{$s \neq s_{\mathrm{goal}}$}
        \State $rhs(s) \gets \displaystyle\min_{s' \in \mathrm{Succ}(s;\sigma^*)}\bigl(g(s') + J(s,s')\bigr)$
      \EndIf
      \If{$g(s)\neq rhs(s)$}
        \State insert $s$ with key $k(s)$ in OPEN
      \EndIf
    \EndFunction

    \Function{OnMapUpdate}{$\mathcal{C}_{\Delta}$}
      \ForAll{edges $e=(u\to v)$ crossing $\mathcal{C}_{\Delta}$}
        \State recompute $J(e)$ \textbf{or} set $J(e)=\infty$ if inadmissible
        \State \Call{UpdateVertex}{$u$}; \ \Call{UpdateVertex}{$v$}
      \EndFor
      \If{committed path cell $>\mathcal{C}_{\text{high}}$ \textbf{ or } $\Delta$path cost $>\epsilon$}
        \State \Call{ComputeShortestPath}{}
      \EndIf
    \EndFunction
  \end{algorithmic}
\end{algorithm}

\begin{figure}[t!]
  \centering
  \begin{subfigure}[b]{0.48\linewidth}
    \includegraphics[width=\linewidth]{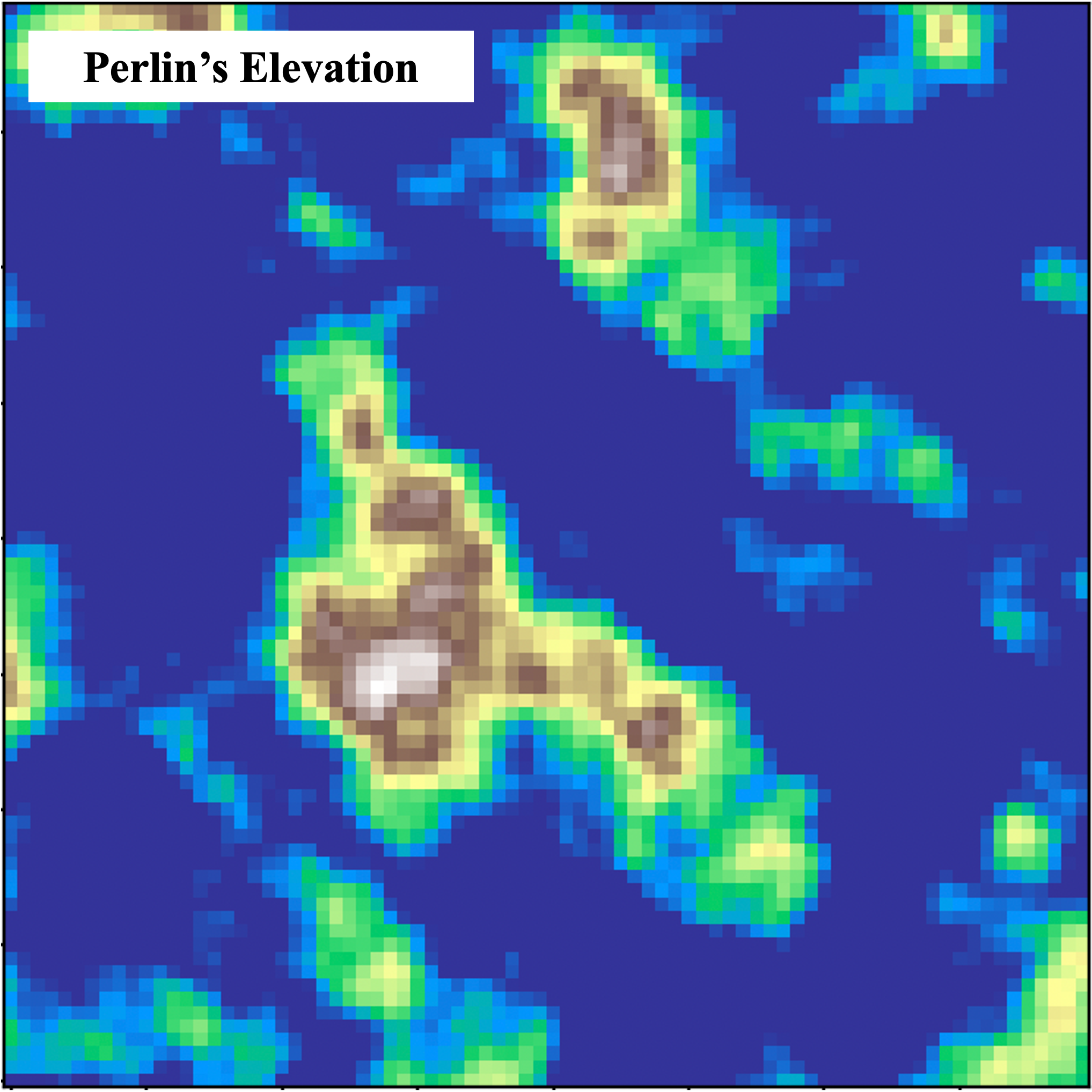}
    \caption{Elevation}
  \end{subfigure}
  \hfill
  \begin{subfigure}[b]{0.48\linewidth}
    \includegraphics[width=\linewidth]{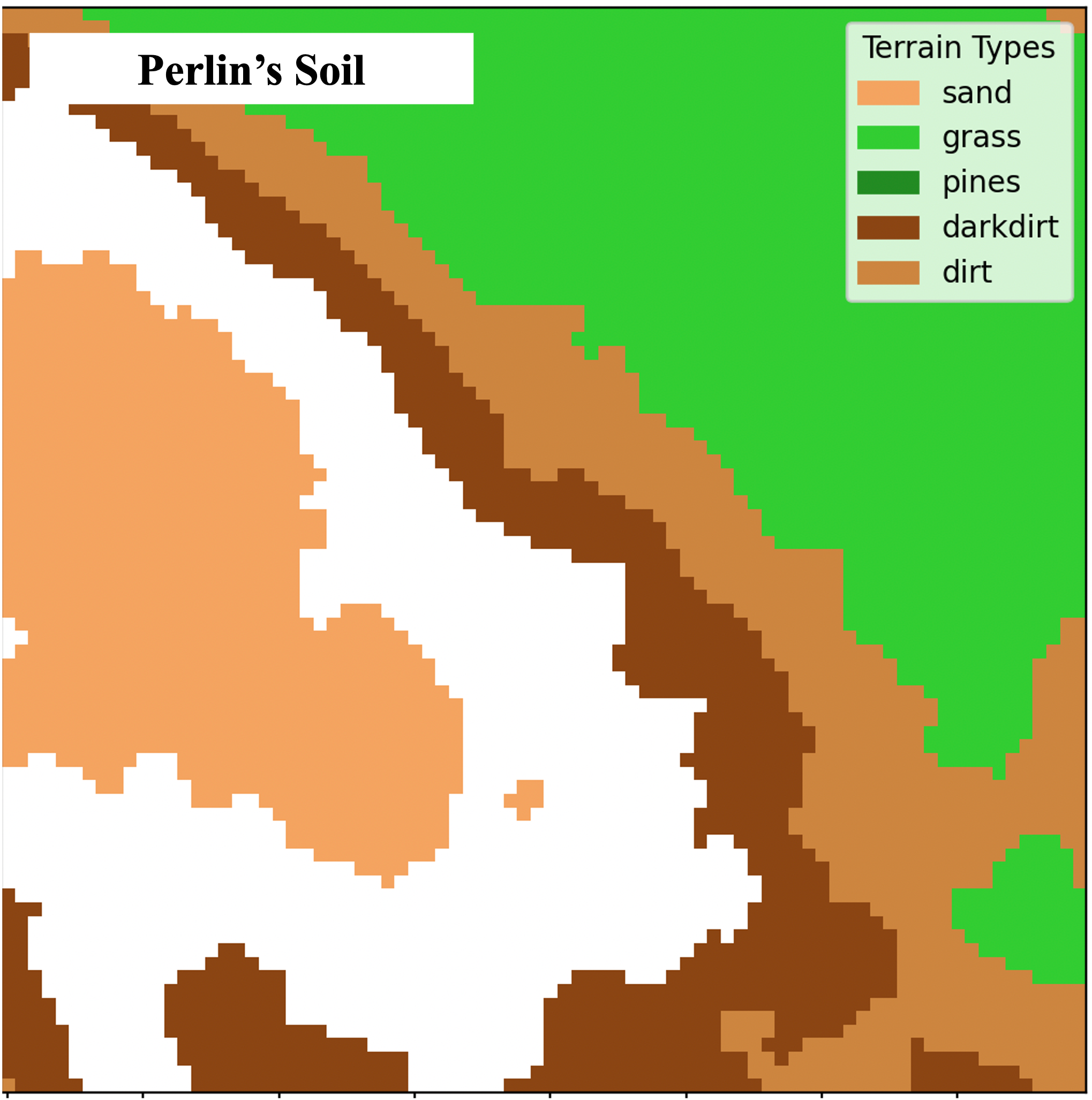}
    \caption{Derived soil mask}
  \end{subfigure}
  \caption{Synthetic test map generated with octave-summed Perlin noise. The same noise field defines both the elevation layer and the soil raster.}
  \label{fig:perlin_demo_results}
\end{figure}

\begin{figure}[t!]
  \centering
  \begin{subfigure}[b]{0.48\linewidth}
    \includegraphics[height=120px]{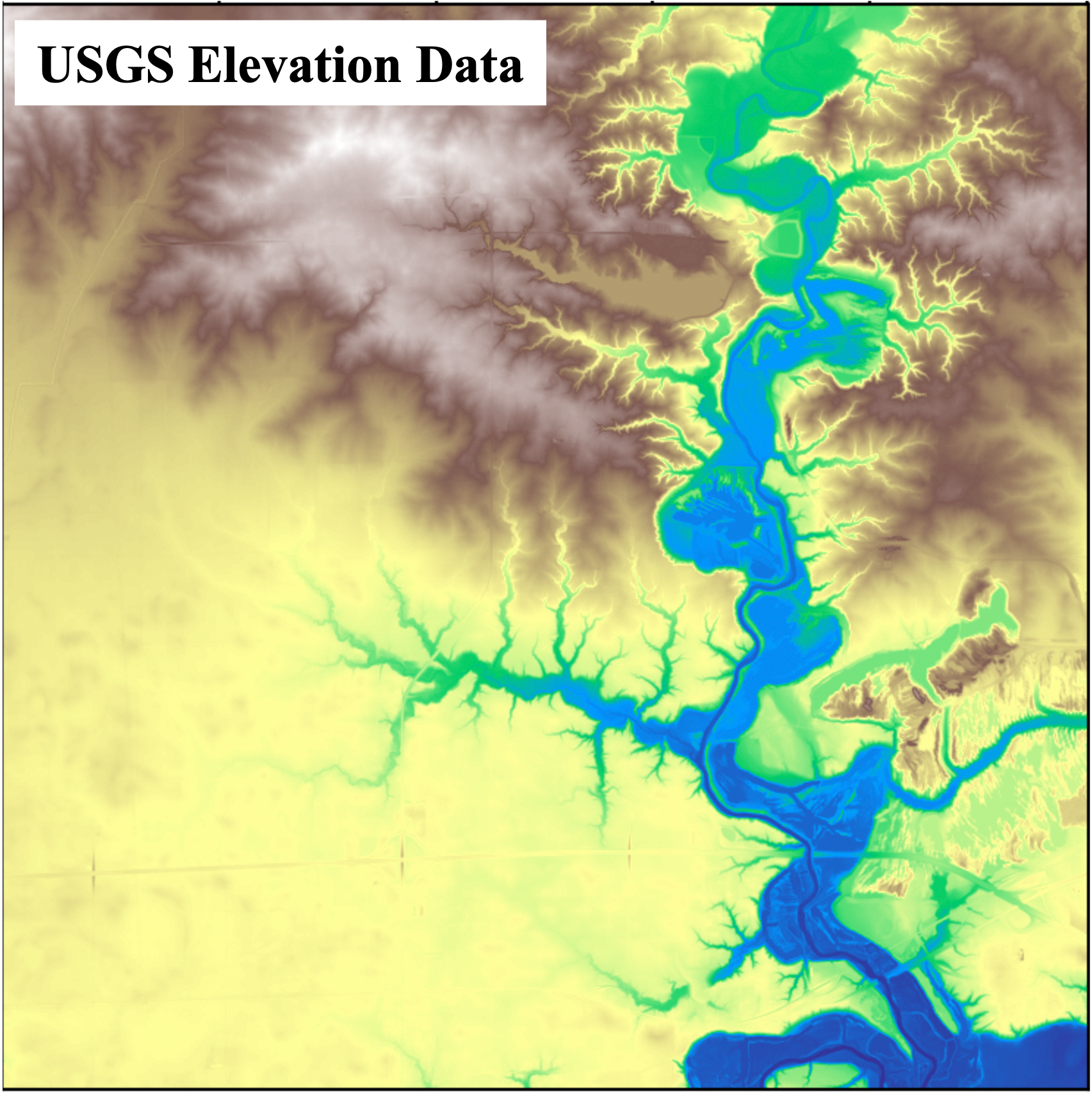}
    \caption{Raw DEM}
  \end{subfigure}
  \hfill
  \begin{subfigure}[b]{0.48\linewidth}
    \includegraphics[height=120px]{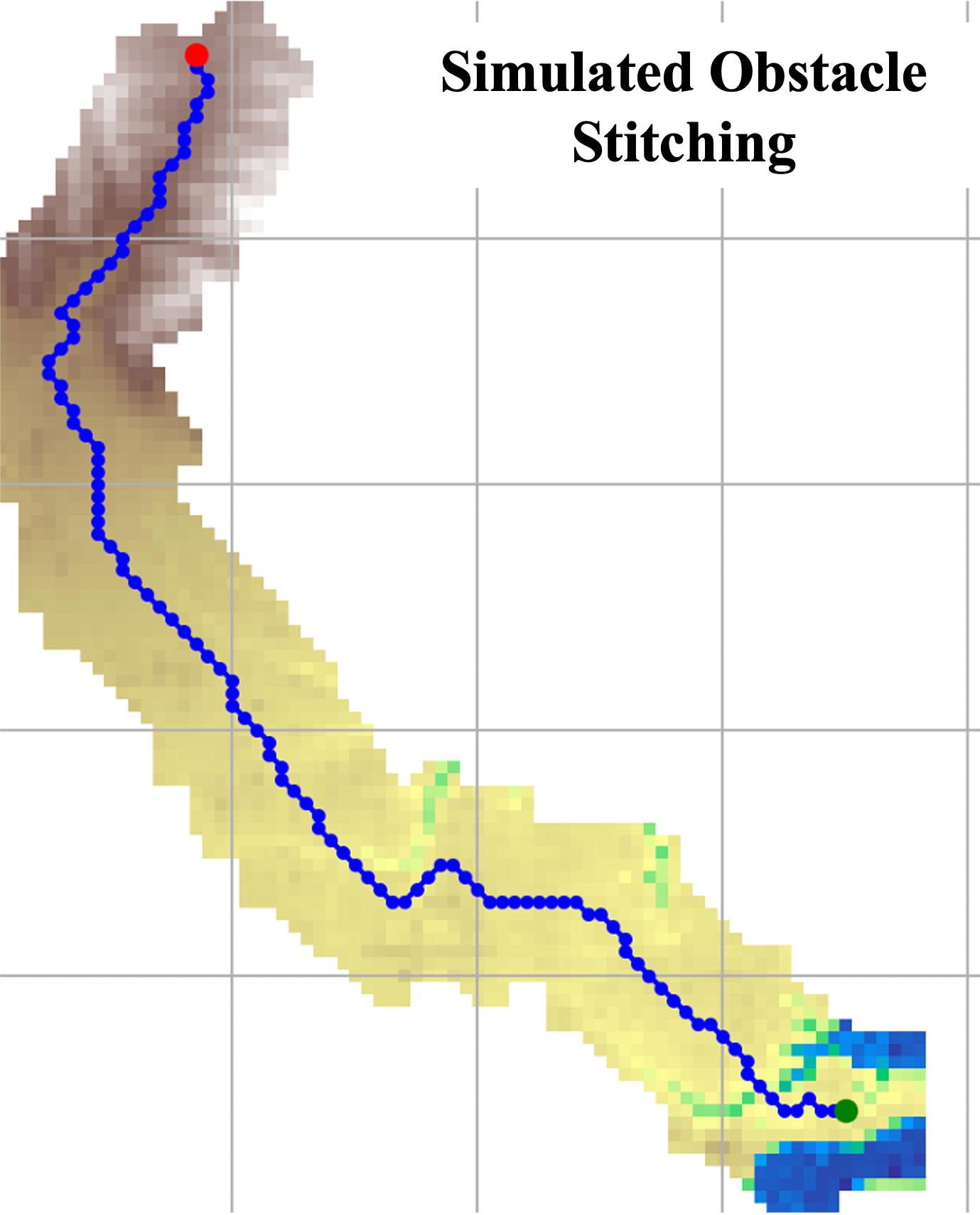}
    \caption{Stitched mosaic}
  \end{subfigure}
  \caption{Map-stitching benchmark: the perception emulator refines a \SI{30}{m} DEM (left) into a high-resolution mosaic (right) as successive LiDAR snapshots are merged.}
  \label{fig:stitch_bench_results}
\end{figure}

\medskip
\noindent\textbf{Perception, stitching, and map memory.}
We evaluate perception robustness with synthetic DEMs generated from octave-summed Perlin noise (Figure~\ref{fig:perlin_demo_results}). Terrain fields are segmented into soil masks to correlate topography and soil type. A perception emulator extracts LiDAR scans (270° FOV, 0.25° resolution), augments them with Gaussian noise and dropouts, and maintains a stitched rolling mosaic (Figure~\ref{fig:stitch_bench_results}). Maps are maintained in a log-odds occupancy grid with octree-based pruning~\cite{wurm_octomap_2010}.

\begin{figure}[t!]
  \centering
  \begin{subfigure}[b]{0.48\linewidth}
    \includegraphics[height=120px]{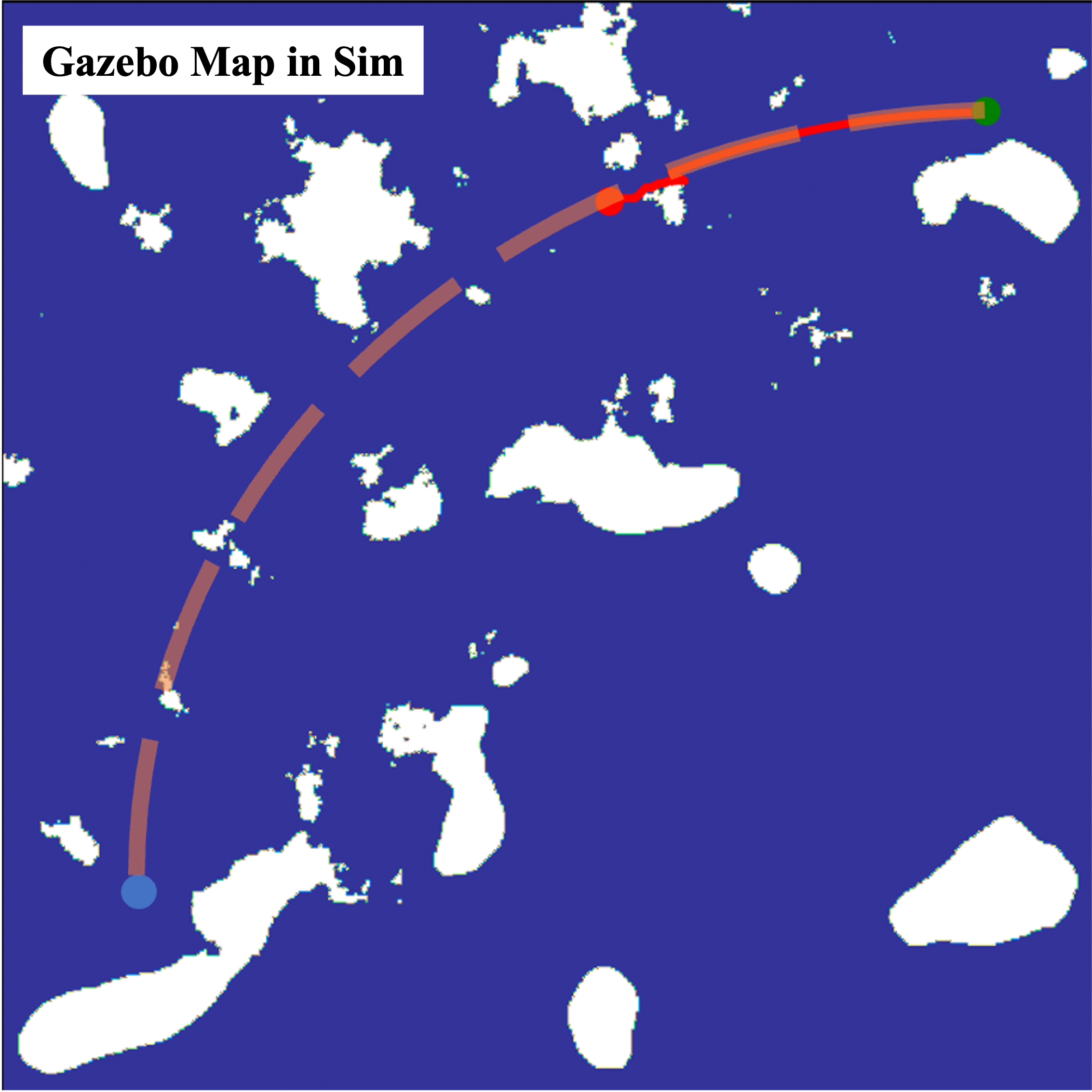}
    \caption{Jackal path through cost map}
    \label{fig:jackal_path}
  \end{subfigure}
  \hfill
  \begin{subfigure}[b]{0.48\linewidth}
    \includegraphics[height=120px]{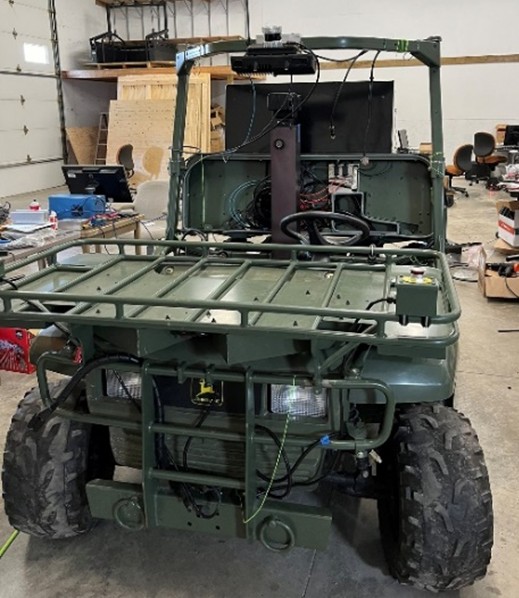}
    \caption{RGator}
    \label{fig:rgator}
  \end{subfigure}
  \caption{Updating cost map with Jackal trajectory (left),  RGator platform with LiDAR and stereo camera kit used in hardware validation experiments (right).}
\end{figure}

\medskip
\noindent\textbf{Computational profile.}
We profile the system in a 100\,m$\times$100\,m simulated environment using a Clearpath Jackal UGV model. The planner runs at 100\,Hz on laptop-class CPUs. Bekker evaluation requires $\approx 0.25$\,ms per tick, while cost fusion and collision checks dominate runtime. Total tick time stays under 10\,ms, enabling real-time use on embedded platforms, as illustrated by the Jackal trajectory in Figure~\ref{fig:jackal_path}.

\medskip
\noindent\textbf{Experimental Setup (Hardware Validation).}
We deployed the planner on the RGator platform~\cite{atlas_r-gator_2004}, equipped with an Ouster OS1 LiDAR, a Carnegie Robotics Multisense S27 stereo camera, a VectorNav VN-310 IMU, and a GPS receiver (Figure~\ref{fig:rgator}). A ruggedized onboard computer with an Intel i7-12700H CPU and 32~GB RAM executed the full pipeline at 100~Hz without GPU acceleration.

To construct the terrain cost field online, we fused depth point clouds from the LiDAR and stereo camera using an extended Kalman filter (EKF) registered against vehicle odometry. The fused cloud was projected into a 2.5D elevation map, defined as a grid map with a single elevation value per cell at 0.1~m resolution. Slope penalties were obtained from finite-difference gradients of this map, while attitude penalties were computed by settling all four wheel contact points onto the DEM surface.

Soil costs were generated using a vision-based terrain identification system~\cite{chen_self-supervised_2023}. Each fused point cloud snapshot was segmented into terrain patches and classified into trained terrain types. These terrain labels indexed into the same lookup table (LUT) of nominal Bekker parameters used in simulation (Table~\ref{tab:bekker_coeffs_verbose}), producing a soil raster aligned with the elevation grid. The raster was then passed through the $C_{\text{soil}}$ formulation described in Section~III-A and fused with slope and attitude penalties to yield the terrain cost map queried by the planner. This ensures that both synthetic and hardware trials rely on a consistent, physics-informed soil model without ad hoc tuning.

As in the simulation pipeline, the offline VD--RRT\(^{*}\) seed path was generated from available DEM data to provide global structure. Since DEMs can be outdated and miss small obstacles, discrepancies were expected; these were handled online by sensor updates and VD--D$^{\ast}$ Lite repairs.

We evaluated the system in outdoor lanes with compact dirt and grassy surfaces over runs of 50--80~m, with elevation changes up to $12^{\circ}$. The planner used the same curvature-feasible lattice as in simulation: Dubins/Reeds--Shepp primitives for differential drive and bicycle arcs for Ackermann steering. Edges were admitted only if all footprint cells remained below the high-risk threshold $C_{high}$, and replans were triggered when a committed path cell violated this bound or when path cost rose beyond $\epsilon$. Success was defined as reaching the goal without collision, immobilization, or exceeding pitch/roll limits.  
In summary, we integrate a Bekker-derived soil cost, slope and attitude penalties, curvature-feasible primitives, and incremental repair into a plug-and-play pipeline for real-time off-road navigation. Each component is physics-grounded yet lightweight, enabling real-time execution on embedded platforms and validated in hardware while remaining planner-agnostic. Building on this methodology, we next evaluate the framework in both large-scale synthetic benchmarks and real-world hardware trials to assess path quality, replanning performance, and robustness.

\section{Results and Analysis}
\label{sec:results}

\medskip
\noindent\textbf{Benchmark setup.}
We benchmark our framework in Gazebo using fifty independent octave-summed Perlin maps of size \(1024\times1024\) cells at \SI{0.1}{m} resolution. Obstacle density is uniformly sampled from \(2\)–\(6\%\). For each map, we evaluate 20 start–goal pairs separated by \(75\!-\!100\)\,\si{\meter}, yielding \(50\times20=1000\) trials per planner. All experiments run on an Intel i7-12700H CPU; the GPU remains idle to reflect the intended RGator embedded deployment.

\medskip
\noindent\textbf{Compared planners.}
We compare four planners: (i) \emph{RRT\(^{*}\) (geom.)}, which steers with Dubins primitives and minimizes Euclidean length while ignoring soil and slope \cite{karaman_sampling-based_2011,reeds_optimal_1990}; (ii) \emph{D\(^{*}\) Lite (grid8)}, which expands an eight-connected lattice with geometric cost \cite{koenig_dlite_2002}; (iii) \emph{RRTx}, which performs incremental rewiring but retains a geometric metric \cite{otte_rrtx_2016}; and (iv) our \emph{hybrid}, which combines VD--RRT\(^{*}\) and VD--D\(^{*}\) Lite atop the Bekker–slope cost and curvature-feasible lattice. A run counts as a failure if any collision, attitude violation, or timeout occurs within the 30\,s wall-clock horizon.

\begin{table}[ht!]
  \centering
  \caption{Synthetic-terrain benchmark on 50 Perlin maps
           (1000 runs, mean $\pm$ SD).}
  \label{tab:synthetic_stats}
  \sisetup{
    separate-uncertainty     = true,
    table-format             = 3.1(2),
    detect-weight            = true,
    detect-inline-family     = text
  }
  \setlength{\tabcolsep}{5pt}
  \footnotesize
  \begin{tabular}{@{}l
                  S
                  S
                  S[table-format=2.0]@{}}
    \toprule
    Planner &
    {Path\,[m]} &
    {CPU\,[ms]} &
    {Succ.\,[\%]} \\ \midrule
    RRT\(^{*}\) (geom.)        & 118.6 \pm 14.3 & 47.2 \pm  9.1 & 64.7 \\
    D\(^{*}\) Lite (grid8)     & 130.5 \pm 18.7 & 22.4 \pm  4.8 & 71.9 \\
    RRTx\,& 121.2 \pm 13.5 & 88.6 \pm 15.4 & 76.1 \\
    \textbf{Hybrid (ours)}
                        & \bfseries 117.9 \pm 12.8
                        & \bfseries 31.6 \pm  6.3
                        & \bfseries 93.4 \\ 
    \bottomrule
  \end{tabular}
\end{table}

\noindent\textbf{Headline results.}
We find that our hybrid planner succeeds in \(93.4\%\) of the trials, exceeding the nearest baseline (RRTx) by more than 17 percentage points. CPU time is cut by one-third relative to RRTx, while path length matches the best geometric planner despite the additional constraints—an expected outcome since curvature-infeasible shortcuts are pruned. Table~\ref{tab:synthetic_stats} summarizes these results.
\begin{figure}[t!]
  \centering
  \includegraphics[width=0.8\linewidth]{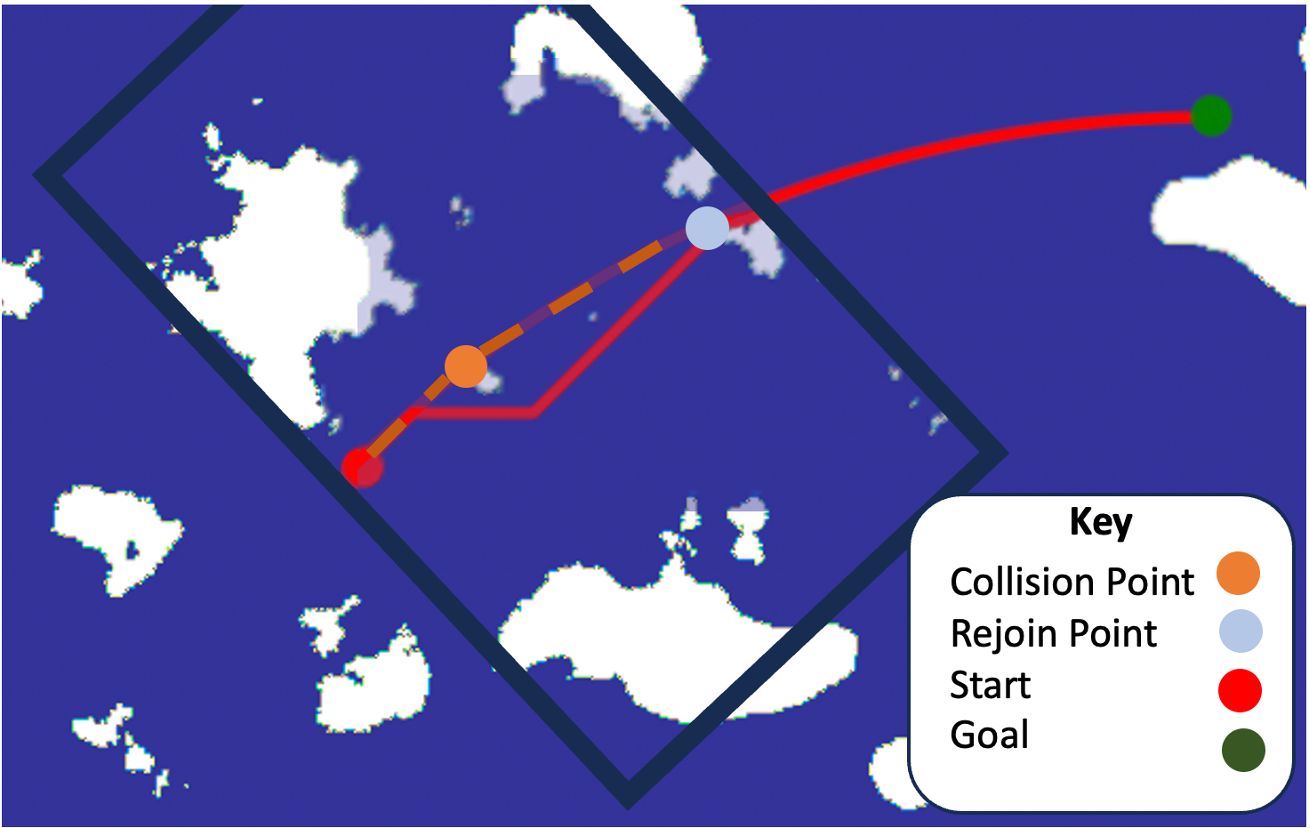}
  \caption{Typical incremental detour injected by VD--D\(^{*}\) Lite when a previously unseen obstacle appears on the current path. The replanner removes only the affected edges, splices a three-vertex repair (violet), and rejoins the original backbone (red) within \SI{2.8}{m}.}
  \label{fig:paths}
\end{figure}

\medskip
\noindent\textbf{Qualitative behavior.}
Figure~\ref{fig:paths} illustrates a representative case: when LiDAR reveals an unseen obstacle, VD–D\(^{*}\) Lite removes only the affected edges and splices in a three-vertex detour before rejoining the original VD–RRT\(^{*}\) backbone. Pure RRT\(^{*}\), on the contrary, must regrow large portions of its tree. This example highlights the key advantage of our hybrid approach: global structure is preserved while local repairs remain fast and bounded.

\medskip
\noindent\textbf{Runtime and robustness.}
We profile the pipeline in Table~\ref{tab:runtime_breakdown}, confirming that it fits comfortably within the 10 ms budget required for a 100 Hz control loop. Cost-map fusion and collision checks dominate runtime, while Bekker evaluation adds only \SI{0.25}{ms} per tick.  

We also test robustness by adding up to \SI{5}{cm} RMS Gaussian noise to the synthetic LiDAR heights. We observe that success rate drops by less than three percentage points, indicating that the attitude penalty helps filter out false obstacles created by noise in the elevation data.
\begin{table}[ht!]
  \centering
  \caption{Runtime per 100 Hz control tick (mean of 2000 ticks).}
  \label{tab:runtime_breakdown}
  \sisetup{table-format=1.2}
  \begin{tabular}{@{}lcc@{}}
    \toprule
    Module & \makebox[3.2cm][c]{Mean\,[ms]} & Share \\ \midrule
    Cost-map fusion (soil + slope)    & 1.17 & 36\% \\
    Edge collision checks             & 0.94 & 29\% \\
    D\(^{*}\) Lite vertex expansions    & 0.61 & 19\% \\
    Dubins / bicycle integration      & 0.28 &  9\% \\
    Bekker-cost evaluation            & 0.25 &  7\% \\ \midrule
    \textbf{Total}                    & \textbf{3.25} & 100\% \\
    \bottomrule
  \end{tabular}
\end{table}

\subsection{Hardware Validation}

We validated the planner on the RGator platform in an outdoor back-lane environment with compact gravel, grass verges, and ad-hoc obstacles. Runs spanned 50--80~m with elevation changes up to $12^{\circ}$. The pipeline was executed on the onboard computer at 100~Hz without GPU acceleration. The RGator platform with the assembled LiDAR + stereo kit is shown in Figure~\ref{fig:rgator}.

\begin{figure}[t!]
  \centering
  \includegraphics[width=0.8\linewidth]{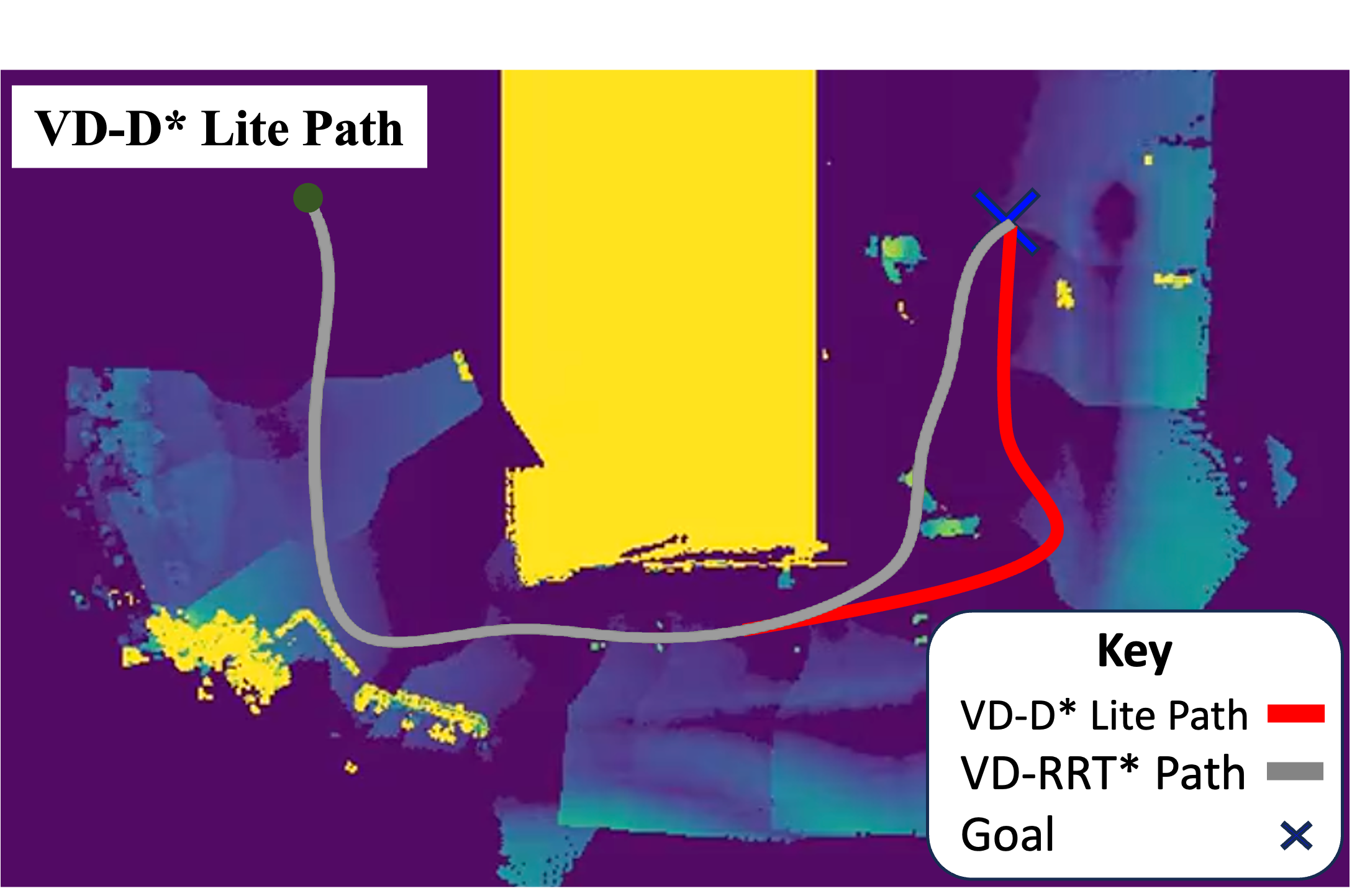}
  \caption{Live VD--D$^{\ast}$ Lite repair (red) over the VD--RRT\(^{*}\) seed (gray) during a hardware run. Yellow cells are impassable; goal is in blue.}
  \label{fig:dstar_live}
\end{figure}

\medskip
\noindent\textbf{Representative run.}
Figure~\ref{fig:dstar_live} shows a typical traverse. The offline VD--RRT\(^{*}\) seed (gray), generated from DEM priors, provided global structure but occasionally mismatched the live environment when small obstacles or terrain changes were not present in the DEM. These discrepancies triggered VD--D$^{\ast}$ Lite, which inserted local repairs (red) based on live sensor updates. Repairs were completed in $<15$~ms, and the vehicle rejoined the original backbone within 3~m, confirming real-time replanning performance consistent with simulation.

\begin{figure}[t]
  \centering
  \includegraphics[width=0.65\linewidth]{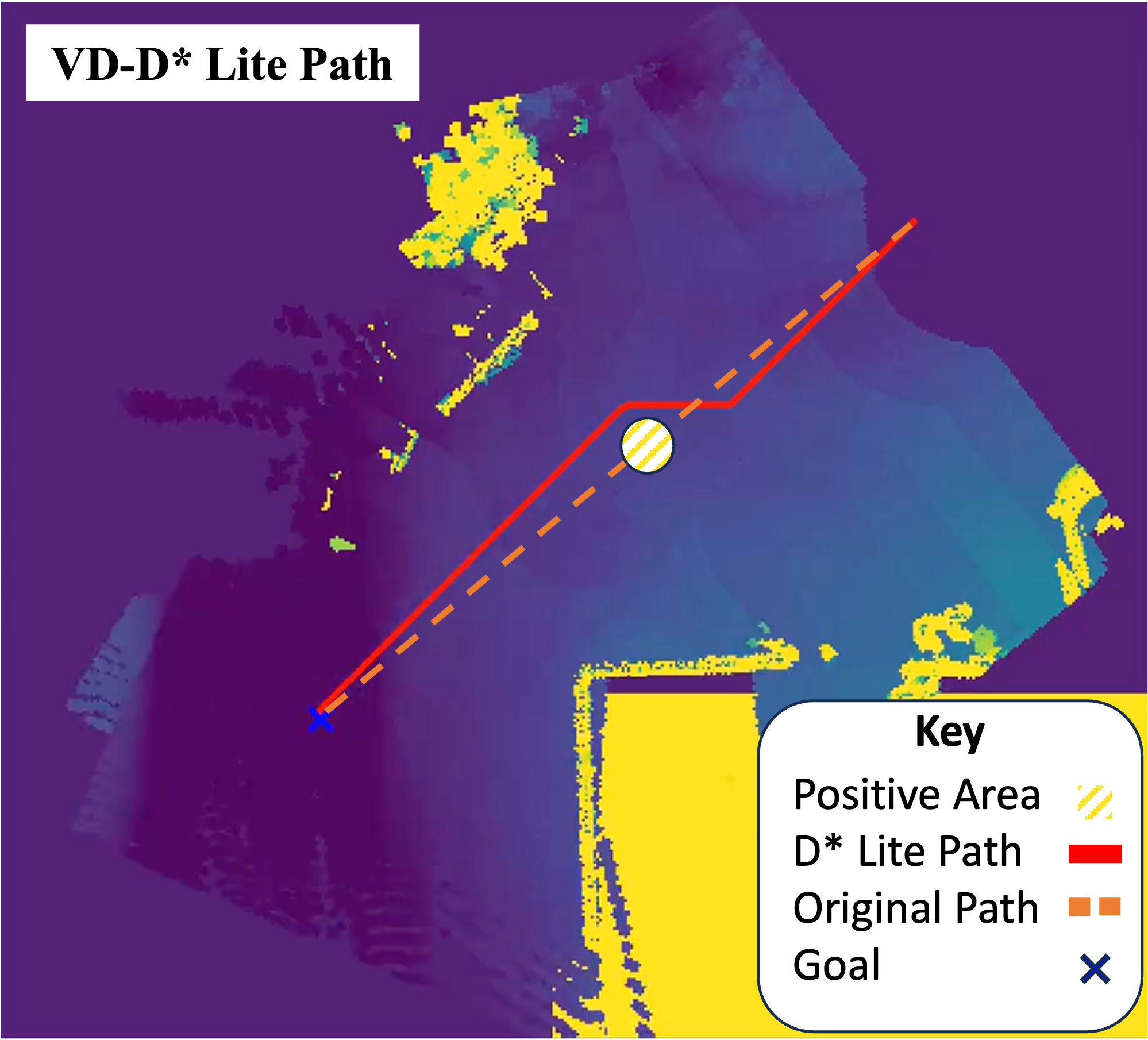}
  \caption{VD--D$^{\ast}$ Lite detour around a 10~cm post (yellow). The original seed path (dashed orange) was rerouted to a curvature-feasible repair (solid red).}
  \label{fig:dstar_pole}
\end{figure}

\medskip
\noindent\textbf{Avoiding narrow obstacles.}
A dedicated trial introduced a single 10~cm post mid-route. As shown in Figure~\ref{fig:dstar_pole}, the post triggered a local repair when the intersecting cell exceeded $C_{high}$. VD--D$^{\ast}$ Lite spliced in a three-vertex detour of 1.8~m radius, clearing the post by 0.62~m laterally. Repair latency was 11~ms, half of the 20~ms cycle budget.

\begin{figure}[t!]
\centering
\begin{subfigure}[b]{\linewidth}
  \centering
  \includegraphics[width=0.64\linewidth]{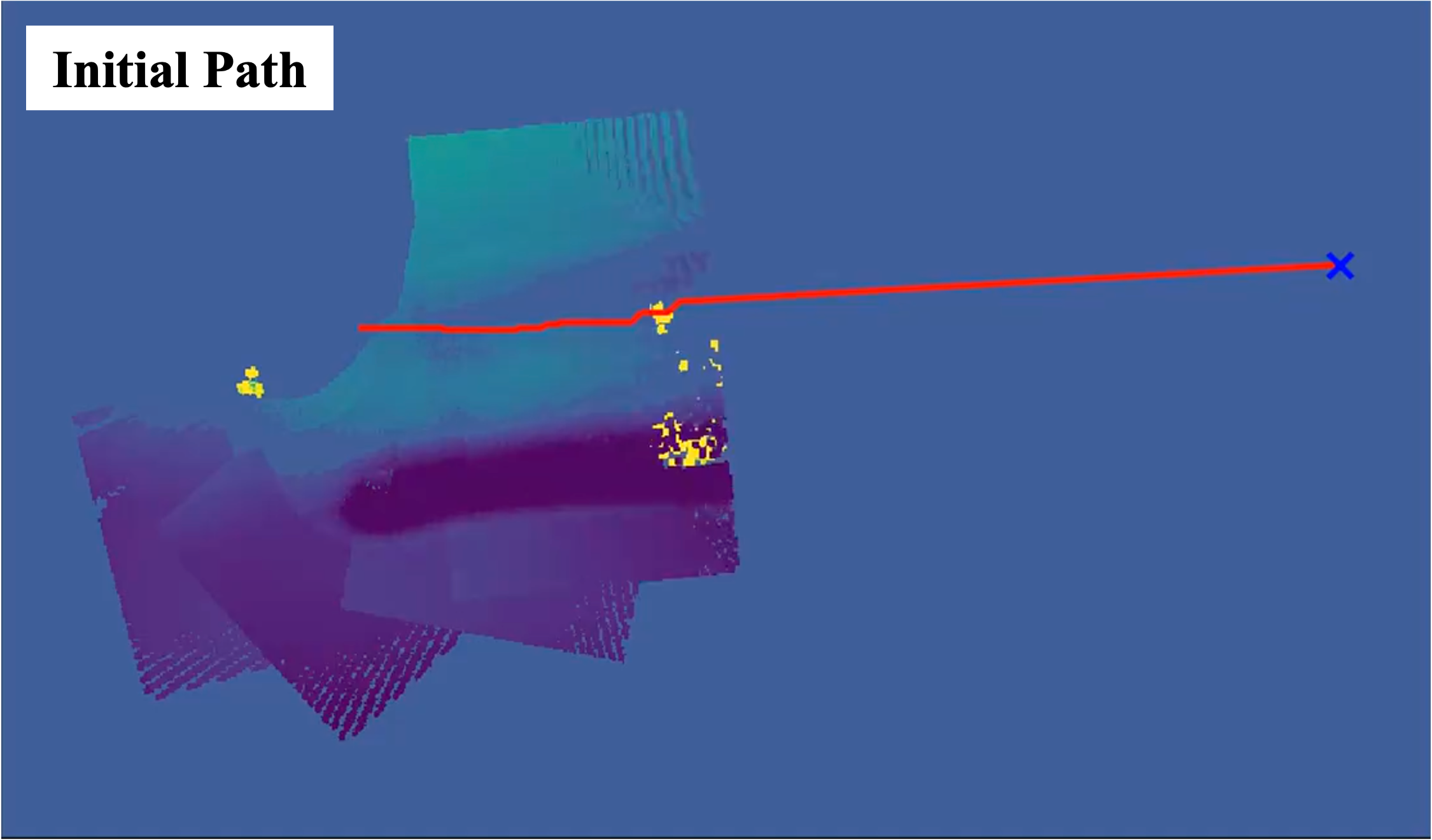}
  \caption{VD--RRT\(^{*}\) seed}\label{fig:rrt_fence}
\end{subfigure}
\begin{subfigure}[b]{0.64\linewidth}
  \centering
  \includegraphics[width=\linewidth]{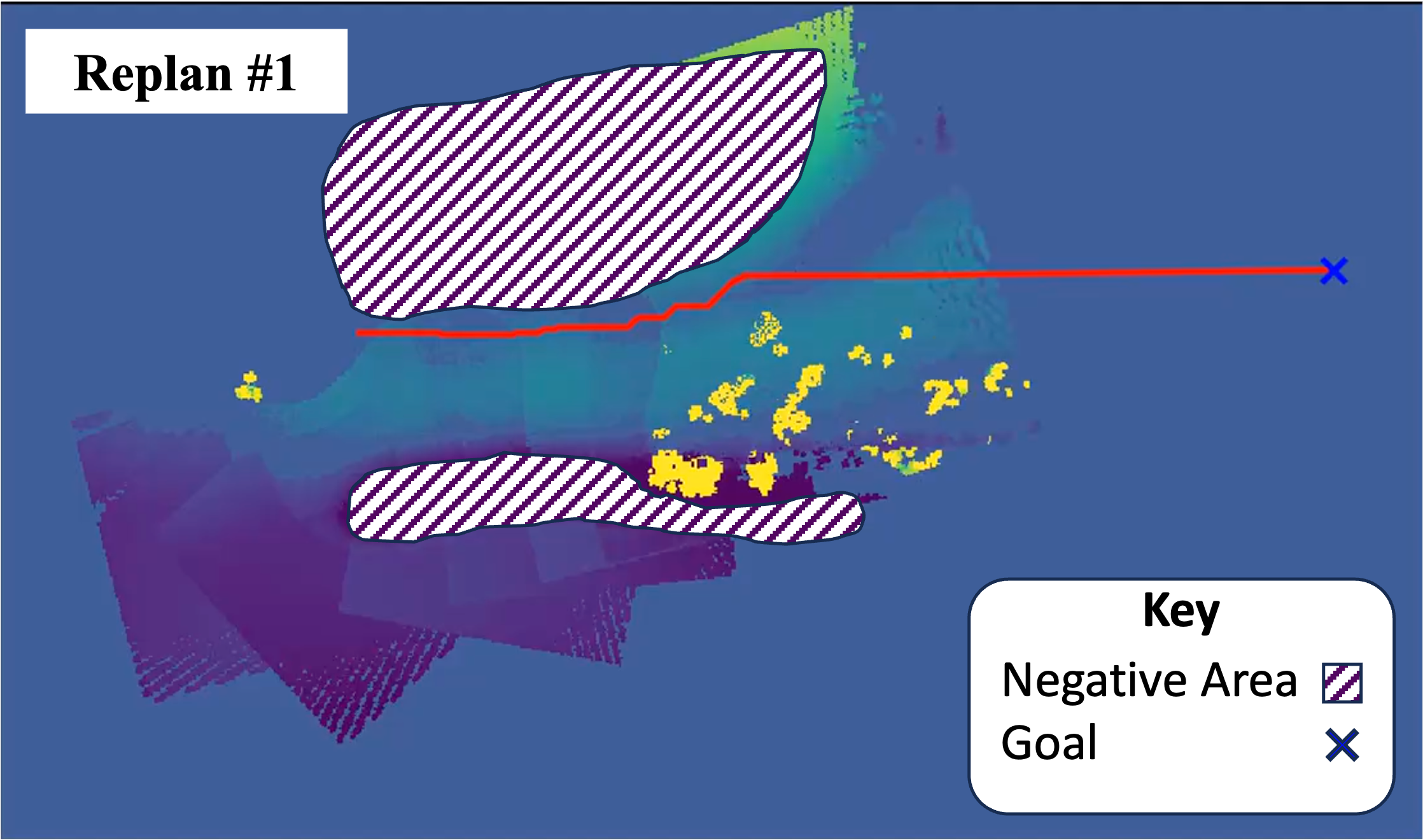}
  \caption{VD--D$^{\ast}$ Lite Replan \#1}\label{fig:dstar_fence1}
\end{subfigure}
\begin{subfigure}[b]{0.64\linewidth}
  \centering
  \includegraphics[width=\linewidth]{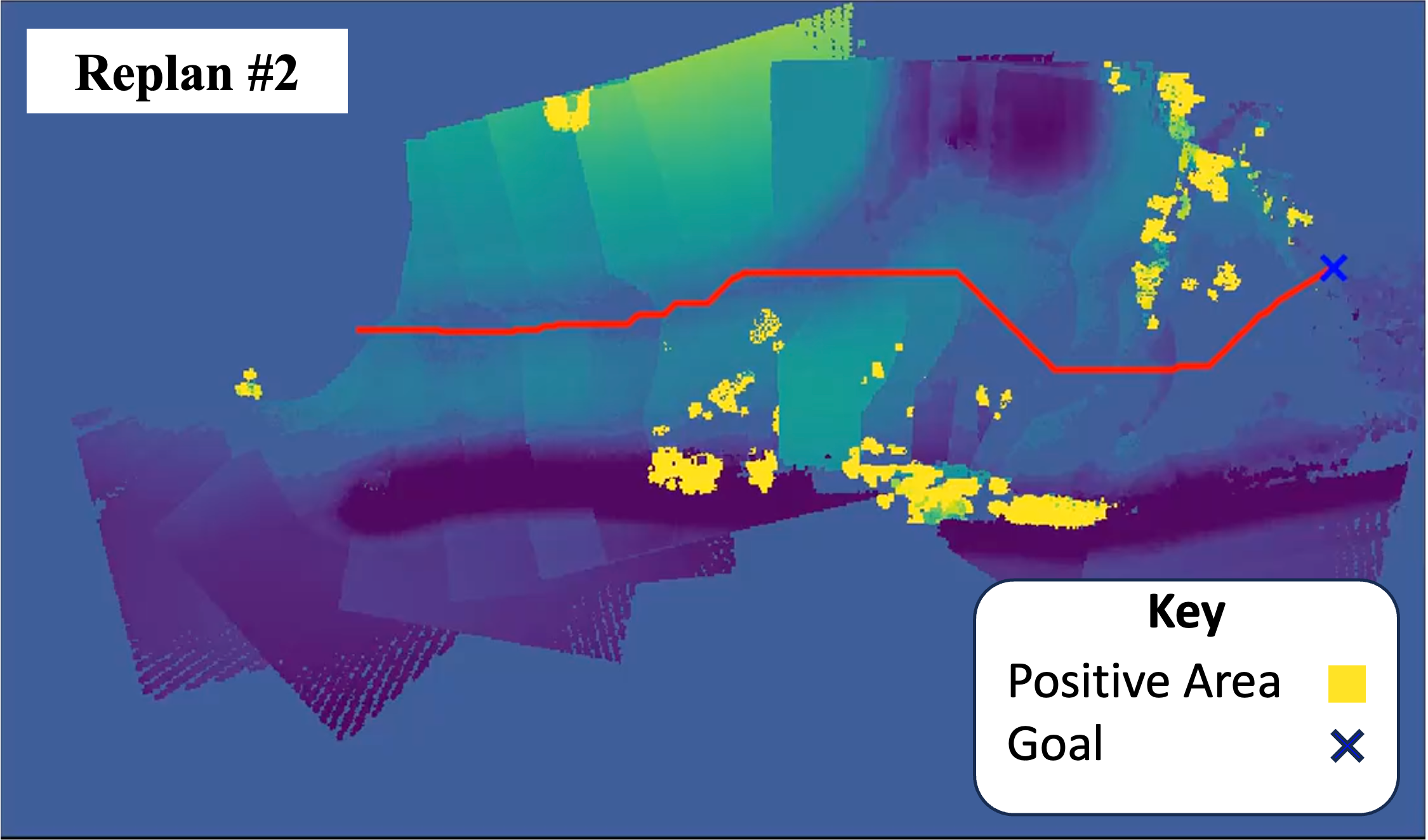}
  \caption{VD--D$^{\ast}$ Lite Replan \#2}\label{fig:dstar_fence2}
\end{subfigure}

\caption{Response to a combined negative-obstacle and fence scenario. (a) shows the VD--RRT\(^{*}\) seed. (b) shows the first VD--D$^{\ast}$ Lite repair when the depression exceeded slope limits. (c) shows the final repair after the fence appears, yielding a safe detour with $\geq0.3$~m clearance.}
\label{fig:neg_obstacle_series}
\end{figure}

\medskip
\noindent\textbf{Skirting a negative obstacle.}
A second vignette combined a shallow depression on one side of the lane with a fence on the other, forcing traversal through a narrow corridor. Figure~\ref{fig:neg_obstacle_series} illustrates the response: the depression exceeded the slope threshold, raising costs to 1, while the fence appeared as a vertical barrier in LiDAR. VD--D$^{\ast}$ Lite injected two successive repairs that yielded a safe detour with $\geq0.4$~m clearance. Latency remained below 15~ms, and the vehicle never paused.

\medskip
\noindent\textbf{Aggregate results.}
Across all hardware trials, the planner consistently reached the goal without collisions or immobilization. Paths averaged $81 \pm 3$~m, closely matching the seed trajectories. Control tick time remained below 10~ms even with frequent repairs ($\sim$5--7 per 100~m), confirming that incremental updates are lightweight. Success rates exceeded 90\%, validating that the physics-aware cost field and curvature-feasible lattice transferred reliably from simulation to hardware.
\begin{table}[t!]
    \centering
    \caption{Hardware trials summary (mean $\pm$ SD).}
    \label{tab:hw_summary}
    \begin{tabular}{lcccc}
        \toprule
        Trial & Path [m] & Tick [ms] & Replans / 100 m & Success [\%] \\
        \midrule
        Lane traverse & $81 \pm 3$ & $8.5 \pm 1.1$ & $\sim$5 & 100 \\
        Narrow post   & $50 \pm 2$ & $9.0 \pm 0.8$ & $\sim$6 & 100 \\
        Neg. obstacle & $65 \pm 4$ & $9.3 \pm 0.9$ & $\sim$7 & 90 \\
        All runs      & --         & $<10$         & $\sim$6 & $>90$ \\
        \bottomrule
    \end{tabular}
\end{table}

These trials demonstrate that the physics-aware cost field and curvature-feasible lattice translate from simulation to real-world deployments. The system handled static hazards, pop-up posts, and combined negative/positive obstacles in real time, confirming that the design remains lightweight, reproducible, and deployable in hardware.

\medskip
\noindent\textbf{Summary.}
Together, these results show that our physics-grounded cost metric and curvature-feasible lattice enable robust real-time navigation. The hybrid VD--RRT\(^{*}\) + VD--D$^{*}$ Lite pipeline achieves near-optimal path length in simulation, millisecond-scale replanning under online map updates, and maintains high robustness when transferred to hardware trials on unstructured terrain. By combining soft-soil awareness, slope and attitude penalties, and curvature-feasible primitives, the framework consistently outperforms purely grid-based or purely sampling-based baselines while remaining lightweight enough for embedded execution.

\section{Discussion}
\label{sec:Discussion}
The synthetic benchmarks confirm that coupling a global, low-dispersion explorer with a curvature-aware incremental repairer combines the strengths of both paradigms. VD--RRT\(^{*}\) provides a globally near-optimal backbone whose edges are guaranteed to satisfy curvature and attitude constraints. VD--D$^{\ast}$ Lite then reuses these edges when local cost updates affect only a small neighborhood, achieving millisecond-scale latency without discarding prior search effort. Although the hybrid incurs a $\sim$9\,ms tick budget compared with 4--5\,ms for pure grid-based planners, this additional cost yields a 17--29 percentage point improvement in success rate and eliminates the need for hand-tuned inflation radii that rigid grids require to avoid soft soil. Runtime profiling confirms that the analytic Bekker term contributes less than 0.3\,ms per tick, validating its use even on embedded CPUs.

Hardware trials reinforce these findings. The DEM-based VD--RRT\(^{*}\) seed paths occasionally mismatched the live environment, particularly when small posts or depressions were absent from the prior map. As expected, this produced multiple repairs (5--7 per 100~m), but tick time remained below 10~ms and every run completed safely, confirming that frequent replans are handled gracefully rather than destabilizing the system. The soil classification LUT likewise generalized across compact dirt, loam, and grass; failures on untrained terrain types remain rare but possible, motivating broader datasets.

Two limitations remain. First, the single-wheel pressure--sinkage model underestimates drawbar pull at high slip. Incorporating shear deformation terms or terradynamic neural surrogates could improve accuracy in sand traps and similar environments. Second, the current pipeline assumes bounded pose error. Coupling edge costs with SLAM uncertainty would allow risk-aware trajectories in GPS-denied settings. A third limitation is that soil classification relies on discrete lookups; integrating continuous online parameter estimation would allow smoother adaptation to unseen terrain. Despite these limitations, the pipeline bridges the gap between physics-agnostic grid planners and expensive terramechanics simulators, delivering safe and near-optimal motion on real vehicles under variable terrain conditions.

\section{Conclusion and Future Work}
\label{sec:conclusion}
We introduced a physics-aware hybrid planner that fuses Bekker-derived soil cost, slope penalties from elevation gradients, and exact Dubins/Reeds--Shepp/bicycle primitives into a single curvature-feasible lattice. Vehicle-Dynamics RRT\(^{*}\) explores this lattice globally, while Vehicle-Dynamics D$^{\ast}$ Lite repairs it incrementally. In simulation, the framework achieved greater than 93\% success on 1000 synthetic off-road trials while meeting a 100\,Hz control budget on laptop-class hardware. Hardware validation on the RGator confirmed that these performance gains transfer to the field: DEM-seeded paths often mismatched live terrain, but the incremental repair loop maintained sub-10\,ms tick times, frequent but lightweight replans, and over 90\% success across outdoor runs. These results demonstrate that physics-grounded costs and curvature-feasible primitives extend beyond controlled benchmarks to real vehicles under variable soil and slope.

\medskip
\noindent\textbf{Future work.} Several directions can extend robustness and broaden applicability:  
\begin{itemize}
  \item \textbf{Ablation and sensitivity studies}: Systematic evaluation of each cost term (soil, slope, attitude) would clarify their individual contributions and guide parameter tuning across environments.  
  \item \textbf{Adaptive soil costs}: Rather than relying on fixed LUT entries, future versions could refine coefficients online by measuring wheel–soil interaction, using learning-based surrogates to adapt to local conditions.  
  \item \textbf{Shear-aware terramechanics}: Incorporating Janosi–Hanamoto shear terms or lightweight surrogates would better capture high-slip behavior in loose soils without breaking the 100\,Hz budget.  
  \item \textbf{Vegetation and brush penalties}: Extending the cost field to model dense vegetation would allow paths that respect vehicle-specific capabilities and avoid mobility losses.  
  \item \textbf{Cross-platform validation}: Deploying the planner on diverse platforms—differential, Ackermann, and tracked—will demonstrate generality beyond the RGator and expose new platform-specific constraints.  
\end{itemize}

\medskip
By unifying terramechanics and non-holonomic motion in a single reusable framework, validating it in both simulation and hardware, and extending it with shear modeling, risk-aware costs, dynamic-agent handling, and larger-scale trials, we move toward reliable real-time autonomy in demanding off-road environments.

\bibliographystyle{IEEEtran}
\bibliography{references}

\end{document}